\documentclass{article} 
\usepackage{iclr2027_conference,times}

\usepackage{amsmath,amsfonts,bm}

\def\eqref#1{equation~\ref{#1}}

\def\1{\bm{1}}

\DeclareMathAlphabet{\mathsfit}{\encodingdefault}{\sfdefault}{m}{sl}
\SetMathAlphabet{\mathsfit}{bold}{\encodingdefault}{\sfdefault}{bx}{n}

\usepackage{hyperref}
\usepackage{url}
\usepackage{wrapfig}
\usepackage{graphicx}
\usepackage{subfig}
\usepackage{algorithmic}
\usepackage{booktabs}
\usepackage{multirow}
\usepackage{caption}
\usepackage[table]{xcolor}
\definecolor{ourcyan}{RGB}{232,245,244}
\newcommand{\ourcell}[1]{\cellcolor{ourcyan}#1}
\usepackage{capt-of}
\newcolumntype{C}[1]{>{\centering\arraybackslash}m{#1}}

\title{Quant\textit{WM}: Temporally Consistent 2-Bit KV Cache Quantization for Video World Models}

\author{
Jiaqi Zhao$^{1,2}$\thanks{Equal contribution.} \thanks{This work was completed while Jiaqi Zhao serves as a visiting student at National University of Singapore.},
Xiaobin Hu$^{2}$\footnotemark[1],
Bo Yin$^{2}$,
Junpeng Jiang$^{1}$,
Miao Zhang$^{1}$\thanks{Corresponding authors.},
Shuicheng Yan$^{2}$
\\[4pt]
$^{1}$Harbin Institute of Technology (Shenzhen), Shenzhen, China \\
$^{2}$National University of Singapore, Singapore
}
\iclrfinalcopy 
\begin{document}

\maketitle

\begin{abstract}
Video generation based world models achieve long-range temporal consistency by storing KV cache during generation, but the continuously growing cache makes KV cache memory become a major deployment bottleneck, which motivates low-bit quantization study for efficiency. Existing 2-bit KV cache quantization methods can achieve nearly quantitatively lossless performance on conventional video benchmarks such as VBench, however, when applied to video world models, we find that they still cause severe temporal flickering and visual degradation. Meanwhile, deeper investigates show that Key quantization produces smaller reconstruction errors than Value, but surprisingly leads to much larger output degradation. We trace this discrepancy to attention in video world models: Key perturbations can change the attention logits, \textit{i.e.}, \textit{QK$^\top$}, and shift the temporal-spatial tokens selected by Queries. These observations motivate us to explicitly preserve attention logits and temporal-spatial token selection during KV cache quantization to alleviate the visual degradation problem. To address this issue, we present \textbf{Quant\textit{WM}}, a training-free 2-bit KV cache quantization framework for video world models. \textbf{Quant\textit{WM}} introduces two complementary techniques to mitigate the attention shifts. Firstly, \textit{quantization-sensitivity-aware clustering (QSAC)} jointly considers historical Query sensitivity and residual ranges to select INT2-friendly Key centroids, which reduces quantization errors in channels that are more critical to attention. In addition, \textit{principal-subspace attention compensation (PSAC)} restores the remaining Key errors along the dominant Query subspace using low-rank projections, which provides a direct and efficient correction to stabilize attention logits. Extensive experiments on LingBot-World-v2, HY-World 1.5, Matrix-Game-2, Longcat-Video and Causal-Forcing demonstrate that \textbf{Quant\textit{WM}} significantly improves visual quality and temporal consistency, while outperforming existing methods across image and video quality metrics with up to \(6.20\times\) KV cache memory compression and limited additional overhead. The project page is at \url{https://quantwm-project.github.io/QuantWM/}.
\end{abstract}

\section{Introduction}
Recent advances in video generation \citep{weissenborn2019scaling,singer2022make,zheng2024open,yang2025cogvideox} have achieved increasingly realistic and temporally consistent visual synthesis, while video world models \citep{alonso2024diffusion,xiao2026worldmem,wang2026matrix} further extend these capabilities to interactive world simulation conditioned on user actions and camera trajectories. In causal and autoregressive video world models \citep{huang2026self,liu2026rolling}, KV cache stores historical Key and Value representations, which allows each generated chunk to attend to relevant world context from previous chunks, so that achieving long-range temporal consistency while supporting continuous interaction. 
However, the large number of spatial-temporal tokens makes the KV cache extremely memory-intensive and becomes a major bottleneck for practical deployment. For example, LingBot-World-v2 \citep{gao2026infinite} requires more than 21 GiB of KV cache memory to generate a 5-second video (about 93 frames), which significantly limits deployment on memory-constrained devices.

\begin{figure*}[t]
\centering
\includegraphics[width=5.5in]{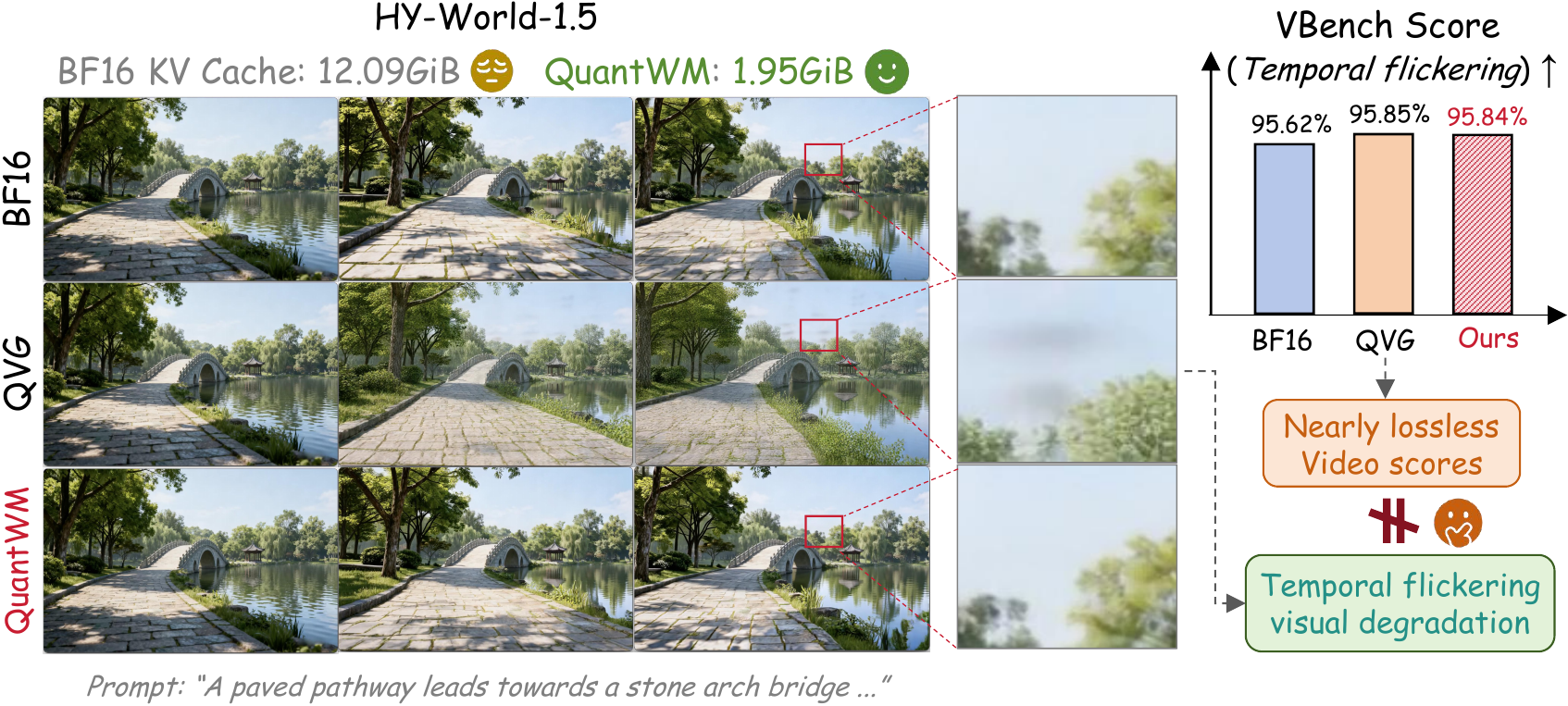}%
\caption{Existing 2-bit KV cache quantization methods achieve nearly lossless video performance but do not fully capture temporal flickering and visual degradation. \textbf{Quant\textit{WM}} effectively improves temporal consistency and significantly reduce the KV cache memory for video world models.}
\label{figure1}
\end{figure*}

Quantization \citep{frantar2022gptq,lin2024awq,xiao2023smoothquant,zhao2024lrquant} is an effective model compression technique which represents high-precision tensors with low-bit values. For KV caches, quantization directly reduces the storage cost of cached Keys and Values, and has achieved promising results in large language models \mbox{\citep{liu2024kivi,hooper2024kvquant,lin2025qserve}}. Particularly, recent methods can compress KV caches to 2-bit or lower with limited performance degradation \citep{zhang2024kv,li2025commvq}. Building on these advances, Quant-VideoGen (QVG) \citep{xi2026quant} extends 2-bit KV cache quantization to autoregressive video generation and shows considerable memory savings with nearly quantitatively lossless performance on video benchmarks, such as VBench \citep{huang2024vbench}.

Although previous methods achieve strong video benchmark performance, we discover an overlooked failure of 2-bit KV cache quantization on video world models. As shown in Figure \ref{figure1}, the quantized videos exhibit clear visual degradation compared with BF16, including temporal flickering, blurring, and artifacts, which are not fully captured by video benchmarks even when their reported scores remain nearly unchanged. This unexpected discrepancy motivates us to further investigate where the visual degradation comes from. We first disentangle the effects of Key and Value quantization and find that \textbf{the degradation is mainly caused by Keys}, where quantizing Keys alone leads to much worse visual quality than quantizing Values. More unexpectedly, Keys themselves exhibit smaller quantization errors than Values but cause larger output errors. This phenomenon indicates that the sensitivity of Keys cannot be explained by quantization error magnitude alone, where more potential issues are hidden \citep{tuncer2026quantized}. 

To understand this challenge, we revisit the attention computation in Transformers. Keys determine the matching scores between Queries and cached tokens through $QK^\top$. Therefore, \textbf{Key perturbations can alter attention logits and change the ranking of temporal-spatial tokens, which may cause frequent token-selection shifts}. In video world models, such shifts will redirect Queries to incorrect cached frames or spatial regions, which leads to temporal flickering and visual degradation.

Based on these insights, we propose \textbf{Quant\textit{WM}}, a training-free 2-bit KV cache quantization framework that preserves attention behavior during quantization on video world models. Specifically, we first introduce \textit{quantization-sensitivity-aware clustering (QSAC)}, which builds on the centroid-residual quantization scheme of QVG and reduces the impact of Key quantization errors on attention logits. Instead of selecting centroids according to K-means clustering \citep{mcqueen1967some}, QSAC jointly considers Query sensitivity and the dynamic range of the K residuals to select centroids with smaller INT2 residual quantization impact on $QK^\top$. Secondly, to further correct the attention shifts, we introduce \textit{principal-subspace attention compensation (PSAC)} to directly compensate the attention logits during inference. In detail, PSAC extracts the dominant Query subspace and represents the remaining Key error components along these sensitive directions with low-rank projections.

We conduct extensive experiments on diverse representative video world models, including LingBot-World-v2 \citep{gao2026infinite}, Matrix-Game-2 \citep{he2025matrix} and HY-World 1.5 \citep{sun2025worldplay}, and two video generation models LongCat-Video \citep{team2025longcat} and Causal-Forcing \citep{zhu2026causal}. The results show that \textbf{Quant\textit{WM}} consistently alleviates the temporal flickering and visual degradation, while maintaining strong performance on benchmarks. Meanwhile, \textbf{Quant\textit{WM}} achieves up to $6.20\times$ KV cache memory compression with only limited inference overhead, which demonstrates its effectiveness and efficiency for KV cache quantization on video world models.

\begin{figure*}[t]
\centering
\includegraphics[width=5.5in]{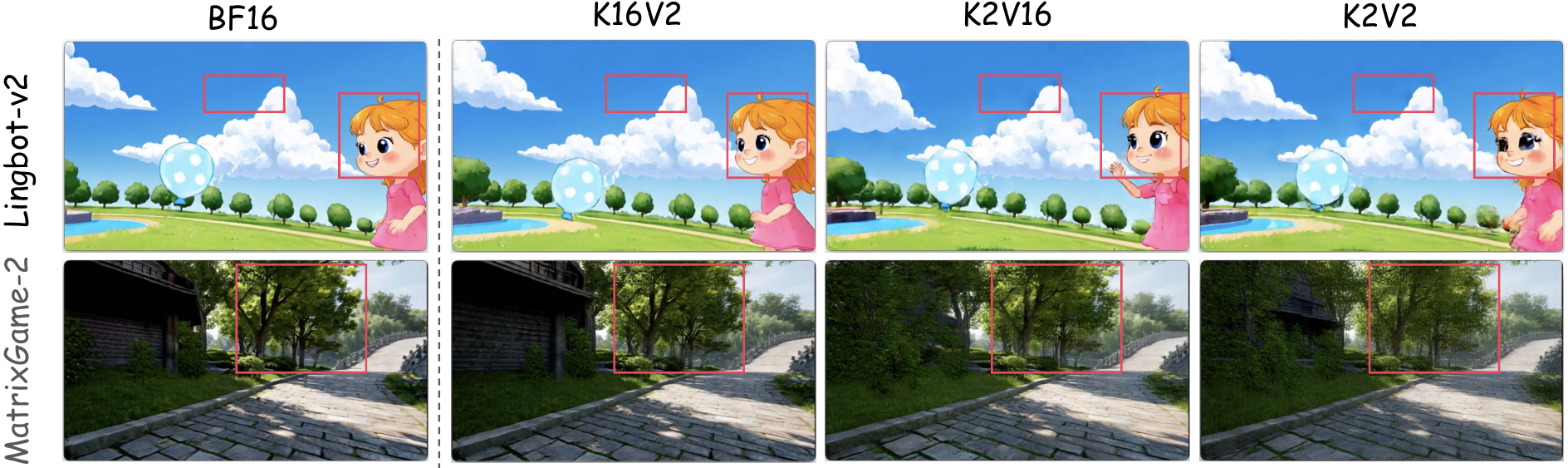}%
\caption{Visual comparison of BF16, K16V2, K2V16 and K2V2 on video world models. As indicated in the \textcolor{red}{red} bounding box area, K16V2 preserves the visual quality well, but K2V16 introduces temporal flickering (above) and visual degradation (below), which indicates Key is more sensitive and dominates the performance degradation.}
\label{figure2}
\end{figure*}

\section{Preliminary}





Quantization is a popular model compression technique which can convert values from full precision to low-bit representations. Given a tensor $X$, its $b$-bit affine quantization and dequantization is formulated as:
\begin{equation}
    X_q = \operatorname{clip}
    \left(
    \left\lfloor \frac{X}{s} \right\rceil + z,
    q_{\min}, q_{\max}
    \right),
    \qquad
    \hat{X}=\mathcal{Q}_b(X)=s(X_q-z),
\end{equation}
where $s$ and $z$ denote the scaling factor and zero-point, respectively, which can be elaborated as:
\begin{equation}
    s = \frac{X_{\max}-X_{\min}}{q_{\max}-q_{\min}},
    \qquad
    z = \left\lfloor q_{\min}-\frac{X_{\min}}{s}\right\rceil.
\end{equation}
The quantization parameters can be shared at different granularities, such as per-token, per-channel, or per-group quantization. For KV cache quantization, we follow the centroid-residual representation proposed by QVG \citep{xi2026quant}. Given a Key token $\mathbf{k}_i\in\mathbb{R}^{d}$, it is assigned to a centroid $\boldsymbol{\mu}_{z_i}$ through K-means clustering \citep{mcqueen1967some} and decomposed as:
\begin{equation}
    \mathbf{r}_i=\mathbf{k}_i-\boldsymbol{\mu}_{z_i}.
\end{equation}
Then the centroid is preserved at BF16 and the residual tensor $\mathbf{r}_i$ is quantized to low-bit, where the reconstructed Key and its quantization error is defined as:
\begin{equation}
    \hat{\mathbf{k}}_i
    =
    \boldsymbol{\mu}_{z_i}
    +
    \mathcal{Q}_b(\mathbf{r}_i).
\end{equation}

\section{Motivation}
\label{motivation}

\subsection{The Overlooked Visual Degradation under 2-Bit KV Quantization}

Existing 2-bit KV cache quantization methods report nearly lossless performance on video benchmarks such as VBench. However, when applied to video world models, we find that these metrics do not fully reflect the degradation introduced by aggressive KV quantization. As shown in Figure \ref{figure1}, compared with the BF16 baseline, the state-of-the-art method QVG produces noticeable temporal flickering, blurring and visual artifacts. These observations suggest that video benchmarks alone may overlook important visual degradation introduced by KV cache quantization. Therefore, we further investigate which component of KV cache is mainly responsible for this degradation.

\begin{figure*}[t]
\centering
\includegraphics[width=5.5in]{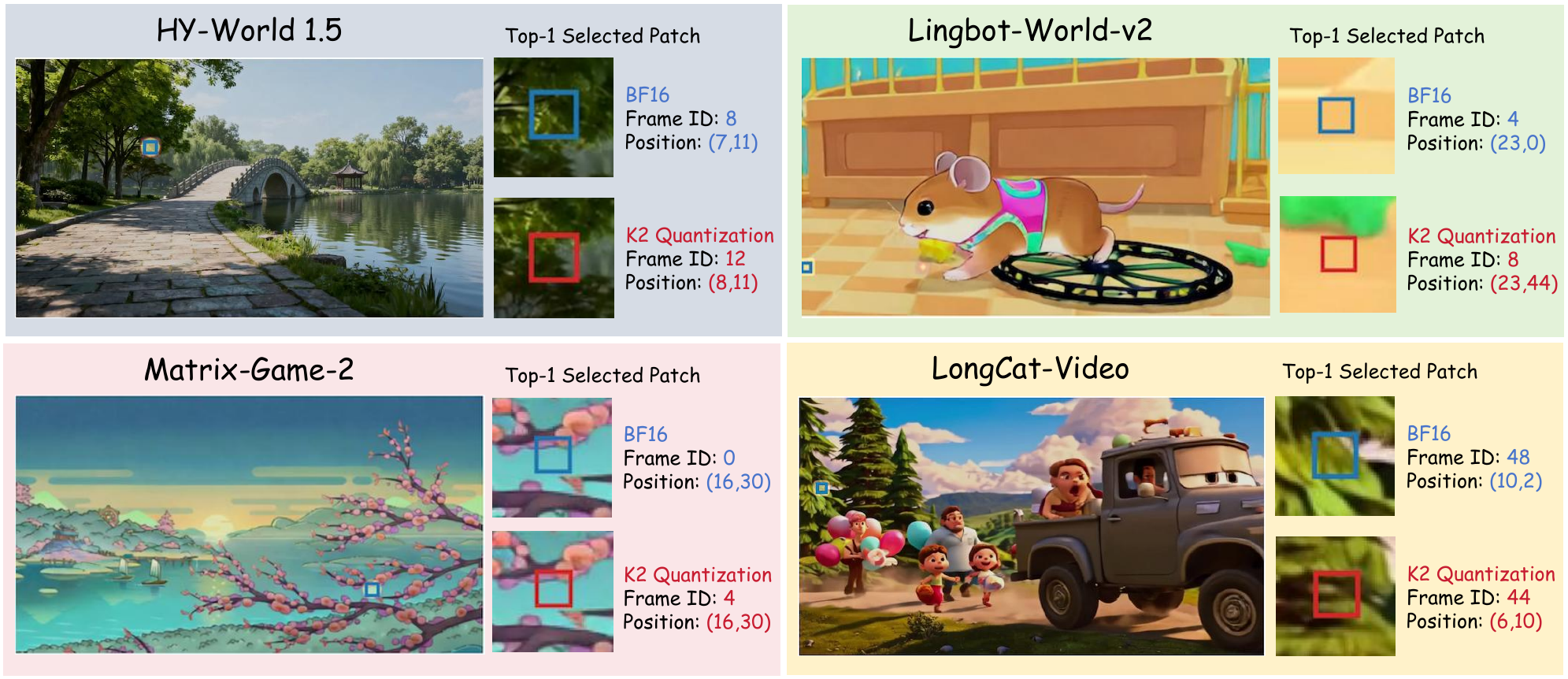}%
\caption{Visualizations of top-1 temporal-spatial token (patch) selection shifts after K quantization under the same query on different video generation and world models.}
\label{figure3}
\end{figure*}

\begin{table}[t]
\centering
\caption{
Quantization error of K2/V2-only and their outputs. Key quantization exhibits smaller reconstruction errors but causes larger output degradation than Value quantization.}
\label{kv_sensitivity}
\small
\begin{tabular}{ll|cc|cc}
\toprule
\textbf{Layer} &
\textbf{Model} &
\cellcolor{blue!7}\textbf{K2 MSE} $\downarrow$ &
\cellcolor{blue!7}\textbf{V2 MSE} $\downarrow$ &
\cellcolor{blue!7}\textbf{K2-Output MSE} $\downarrow$ &
\cellcolor{blue!7}\textbf{V2-Output MSE} $\downarrow$ \\
\midrule

\multirow{3}{*}{Middle}
& Matrix-Game-2
& \textbf{0.1460} & 0.2335
& 0.0853 & \textbf{0.0560} \\
& LingBot-v2
& \textbf{0.2717} & 0.3557
& 0.1868 & 0.1946 \\
& HY-World 1.5
& \textbf{0.1340} & 0.2973
& 0.0821 & \textbf{0.0536} \\
\midrule

\multirow{3}{*}{Last}
& Matrix-Game-2
& \textbf{0.0514} & 0.3005
& 0.3107 & \textbf{0.1860} \\
& LingBot-v2
& \textbf{0.1608} & 0.3008
& 0.2256 & \textbf{0.1039} \\
& HY-World 1.5
& \textbf{0.0090} & 0.2939
& 0.5884 & \textbf{0.1679} \\
\bottomrule
\end{tabular}
\end{table}

\subsection{Key Quantization Dominates Visual Degradation}

\begin{wraptable}{r}{6.4cm}
\vspace{-1.2em}
\centering
\caption{Visual quality of the samples under different setting, where K2-only causes larger visual degradation than V2-only. MG-2 and LB-2 are short for Matrix-Game-2 and Lingbot-World-v2, respectively.
}
\label{kv_bit_ablation}
\small
\setlength{\tabcolsep}{4pt}
\renewcommand{\arraystretch}{1.05}
\begin{tabular}{ll|ccc}
\toprule
\textbf{Model} &
\textbf{Setting} &
\cellcolor{blue!7}\textbf{PSNR} $\uparrow$ &
\cellcolor{blue!7}\textbf{SSIM} $\uparrow$ &
\cellcolor{blue!7}\textbf{LPIPS} $\downarrow$ \\
\midrule

\multirow{3}{*}{MG-2}
& K2V16
& 18.47 & 0.547 & 0.219 \\
& \textbf{K16V2}
& \textbf{21.58} & \textbf{0.708} & \textbf{0.126} \\
& K2V2
& 18.32 & 0.500 & 0.261 \\
\midrule

\multirow{3}{*}{LB-2}
& K2V16
& 15.38 & 0.355 & 0.263 \\
& \textbf{K16V2}
& \textbf{17.70} & \textbf{0.476} & \textbf{0.157} \\
& K2V2
& 15.79 & 0.376 & 0.264 \\
\bottomrule
\end{tabular}
\vspace{-1.5em}
\end{wraptable}

To locate the main source of visual degradation, we separately quantize Key and Value caches using QVG. As shown in Figure \ref{figure2}, K2V16 produces more severe visual degradation than K16V2. On these samples, the same trend is reflected by frame-level quality metrics as shown in Table \ref{kv_bit_ablation}, where K16V2 achieves clearly better PSNR, SSIM and LPIPS than K2V16, while K2V2 performs similarly to K2V16. These results indicate that the degradation of 2-bit KV cache quantization on video world models is mainly dominated by Key quantization.

More surprisingly, as shown in Table \ref{kv_sensitivity}, the reconstruction error of quantized Keys is even smaller than that of Values, while K2V16 generally introduces a larger output error than K16V2. This suggests that the sensitivity of Keys cannot be explained by the magnitude of their quantization error.

\subsection{Key Quantization Perturbs Temporal-Spatial Token Selection}

To understand why Key is quantization-sensitive, we revisit the attention computation. Given Query $\mathbf{Q}$ and Key $\mathbf{K}$, the attention logits and the post-quantization logits perturbation can be elaborated as:
\begin{equation}
    \mathbf{L}=\frac{\mathbf{Q}\mathbf{K}^{\top}}{\sqrt{d}},
    \qquad
    \Delta\mathbf{L}
    =
    \frac{\mathbf{Q}(\mathbf{K}-\hat{\mathbf{K}})^{\top}}{\sqrt{d}}.
\end{equation}
Therefore, even a small Key perturbation may change the relative ordering of attention logits and shift the temporal-spatial tokens selected by Queries.

To validate this analysis, we visualize the temporal-spatial tokens selected by Queries. As shown in Figure \ref{figure3}, 2-bit Key causes clear token-selection shifts compared with BF16, where Queries attend to different historical frames or spatial regions. In video world models, such attention mismatch may redirect the current chunk to inconsistent historical frames or spatial regions, which finally leads to temporal flickering and visual degradation. Accordingly, in addition to reconstruction error, KV cache quantization should also consider to reduce its impact on attention logits.

\section{Quant\textit{WM}}

In this section, we present \textbf{Quant\textit{WM}}, a training-free 2-bit KV cache quantization framework for video world models. We first introduce \textit{quantization-sensitivity-aware clustering (QSAC)} to reduce the impact of Key quantization on attention logits. And then we present \textit{principal-subspace attention compensation (PSAC)} to further correct the attention shifts. 

\subsection{Quantization-Sensitivity-Aware Clustering (QSAC)}
\label{sec:qsac}

Existing centroid-residual KV quantization methods (QVG) typically apply K-means to assign each Key token to its nearest centroid and then quantizing the residual tensor. Such strategy mainly considers the distance between tokens and centroids but ignores the impact of the quantized residual tensor. As discussed in Section~\ref{motivation}, even a small Key quantization error may cause large perturbations to attention logits when it interacts with sensitive Query channels. To address this issue, we introduce QSAC, which selects centroids based on the impact of residual quantization on attention logits. Specifically, it jointly considers Query-channel sensitivity and the dynamic range of residual groups so that it can introduce smaller quantization perturbations to $QK^\top$.

To approximate the Q-channel sensitivity during causal generation, we use the Queries observed in previous generation steps and maintain their second-order statistics for each attention head $h$:
\begin{equation}
    \mathbf{M}_{h}^{t-1}
    =
    \frac{1}{N_{<t}}
    \sum_{\tau<t}\sum_i
    \mathbf{q}_{\tau,i,h}
    \mathbf{q}_{\tau,i,h}^{\top}.
\label{eq6}
\end{equation}
The second-order Query statistics directly follows from the squared attention-logit error introduced by Key quantization. Given a Key quantization error $\mathbf{e}=\mathbf{k}-\hat{\mathbf{k}}$, we have:
\begin{equation}
    \mathbb{E}_{\mathbf{q}}
    \left[(\mathbf{q}^{\top}\mathbf{e})^2\right]
    =
    \mathbf{e}^{\top}\mathbf{M}_{h}^{t-1}\mathbf{e},
\label{eq7}
\end{equation}
where $\mathbf{M}_{h}^{t-1}$ characterizes the sensitivity of different Key-error directions to historical Queries. However, directly using the full matrix $\mathbf{M}_{h}^{t-1}$ for centroid assignment introduces additional computation overheads. To reduce computation, we use its diagonal approximation, where $[\mathbf{M}_{h}^{t-1}]_{cc}=\mathbb{E}[q_c^2]$ weights the contribution of channel $c$ to the expected logit perturbation:
\begin{equation}
\mathbf{e}^{\top}\mathbf{M}_{h}^{t-1}\mathbf{e}
\approx
\sum_c[\mathbf{M}_{h}^{t-1}]_{cc}e_c^2,
\qquad
w_{h,c}
=
\frac{[\mathbf{M}_{h}^{t-1}]_{cc}}
{\frac{1}{d}\sum_{c'}[\mathbf{M}_{h}^{t-1}]_{c'c'}}.
\label{eq11}
\end{equation}
Channels with larger $w_{h,c}$ are more sensitive to Key quantization errors. Based on this sensitivity, QSAC first defines a Query-aware distance:
\begin{equation}
    d_{\mathrm{Q}}(\mathbf{k}_i,\boldsymbol{\mu}_j)
    =
    \sum_c w_{h,c}(k_{i,c}-\mu_{j,c})^2,
\label{eq9}
\end{equation}
which is used to efficiently select a top-$M$ candidate centroids set $\mathcal{C}_i$. Subsequently, we consider the quantization difficulty of the residuals. For each candidate centroid $\boldsymbol{\mu}_j$, we define the residual and its group-wise dynamic range as:
\begin{equation}
    \mathbf{r}_{ij}=\mathbf{k}_i-\boldsymbol{\mu}_j,
    \qquad
    R_g(\mathbf{r}_{ij})
    =
    \max_{c\in\mathcal{G}_g}r_{ij,c}
    -
    \min_{c\in\mathcal{G}_g}r_{ij,c},
\end{equation}
where $\mathcal{G}_g$ denotes the set of channels in the $g$-th group. For uniform $b$-bit quantization, the corresponding scaling factor is
$\Delta_g=R_g/(2^b-1)$. Then according to \citep{widrow1996statistical}, the quantization error can be approximated as:
\begin{equation}
    \mathbb{E}[e_c^2]
    \approx
    \frac{\Delta_g^2}{12}
    =
    \frac{R_g^2}{12(2^b-1)^2}.
\label{quant-equal}
\end{equation}

Since $1/[12(2^b-1)^2]$ is shared by all candidate centroids, we remove it and combine Eq. \ref{quant-equal} with Eq. \ref{eq11} to get the expected squared attention-logit perturbation for a candidate centroid, which we define it as the QSAC score and then use it to select the optimal centroid:
\begin{equation}
    S(\mathbf{k}_i,\boldsymbol{\mu}_j)
    =
    \sum_g
    \left(
        \sum_{c\in\mathcal{G}_g} w_{h,c}
    \right)
    R_g(\mathbf{r}_{ij})^2,
    \qquad
    z_i
    =
    \arg\min_{j\in\mathcal{C}_i}
    S(\mathbf{k}_i,\boldsymbol{\mu}_j).
\label{eq13}
\end{equation}
In this way, QSAC can prioritizes Query-sensitive groups and successfully incorporates the approximated residual quantization error into centroid selection, so that the selected centroid is expected to introduce smaller perturbations to $QK^\top$ after quantization. 

\subsection{Principal-Subspace Attention Compensation (PSAC)}
\label{sec:psac}

\begin{wrapfigure}{r}{2.3in}
    \vspace{-1.5em}
    \centering
    \includegraphics[width=2.3in]{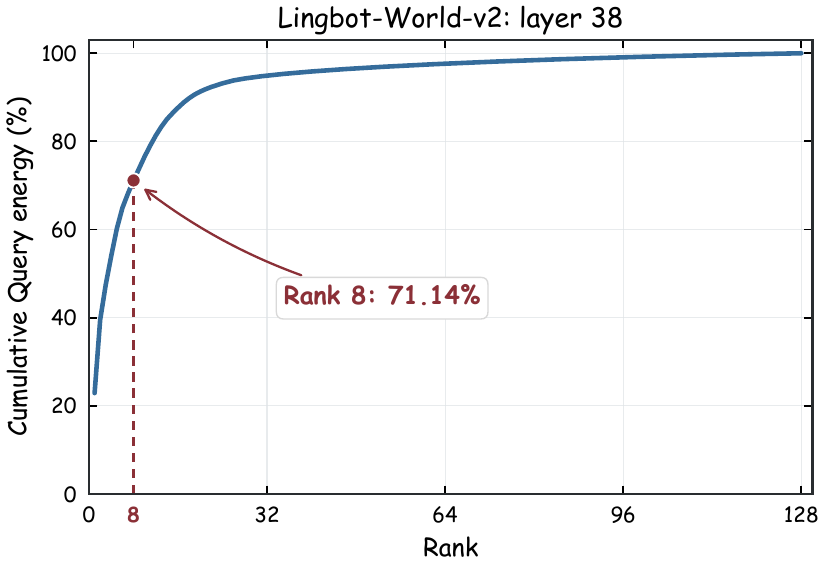}
    \caption{Cumulative Query energy in LingBot-v2, where top-8 directions include most of the total Query energy.}
    \label{rank8}
    \vspace{-1em}
\end{wrapfigure}

Due to the limited representation capacity of 2-bit quantization, Key quantization errors cannot be fully eliminated even after QSAC, which may still cause attention shifts. Therefore, we further introduce PSAC to directly correct the remaining logit perturbations.

Let $\hat{\mathbf{K}}_0$ denote the dequantized Key after QSAC, and we define the remaining quantization error as:
\begin{equation}
    \mathbf{E}
    =
    \mathbf{K}-\hat{\mathbf{K}}_0.
\end{equation}
Reusing the historical Query statistics $\mathbf{M}_{h}^{t-1}$ introduced in Section~\ref{sec:qsac}, the expected logit perturbations caused by $\mathbf{E}$ can be written as:
\begin{equation}
    \mathcal{D}(\mathbf{E})
    =
    \mathbb{E}_{\mathbf{Q}}
    \left[
        \left\|\mathbf{Q}\mathbf{E}^{\top}\right\|_F^2
    \right]
    =
    \operatorname{Tr}
    \left(
        \mathbf{E}\mathbf{M}_{h}^{t-1}\mathbf{E}^{\top}
    \right).
\end{equation}
However, directly storing full quantization error $\mathbf{E}$ for logits compensation would introduce additional memory overheads. Inspired by \citep{liu2025cola,zhao2025boost} which claim that the hidden representations in Transformers often exhibit low-rank structures, we examine the energy distribution of historical Queries and find that the energy is highly concentrated in a few principal directions. As shown in Figure \ref{rank8}, the top-8 directions include more than 60\% of the total Query energy, which suggest that we can focus on the error components along these dominant Query directions. Specifically, we perform eigenvalue decomposition for $\mathbf{M}_{h}^{t-1}$:
\begin{equation}
    \mathbf{M}_{h}^{t-1}
    =
    \mathbf{U}_h
    \boldsymbol{\Lambda}_h
    \mathbf{U}_h^{\top},
    \qquad
    \lambda_1 \geq \lambda_2 \geq \cdots \geq \lambda_d.
\end{equation}
Each eigenvalue represents the historical Query energy along its corresponding eigenvector. Therefore, Key quantization errors projected onto directions with larger eigenvalues have a larger expected impact on attention logits. We preserve the top-$r$ eigenvectors $\mathbf{U}_r$ as the principal Query subspace and restore the remaining Key quantization error along these directions:
\begin{equation}
    \mathbf{C}
    =
    \mathbf{E}\mathbf{U}_r,
\end{equation}
where we preserve both $\mathbf{C}$ and $\mathbf{U}_r$ during inference. Finally, we compensate these error components after quantization, where the attention logits will be corrected as follows:
\begin{equation}
    \hat{\mathbf{K}}
    =
    \hat{\mathbf{K}}_0
    +
    \mathbf{C}\mathbf{U}_r^{\top},
    \qquad
    \mathbf{Q}\hat{\mathbf{K}}^{\top}
    =
    \mathbf{Q}\hat{\mathbf{K}}_0^{\top}
    +
    \mathbf{Q}\mathbf{U}_r\mathbf{C}^{\top}.
\end{equation}
In our implementation, we set $r=8$. Since both $\mathbf{C}$ and $\mathbf{U}_r$ are low-rank, PSAC can directly compensates the attention-logit shift without extensive extra overheads.

\begin{figure*}[t]
\centering
\includegraphics[width=5.5in]{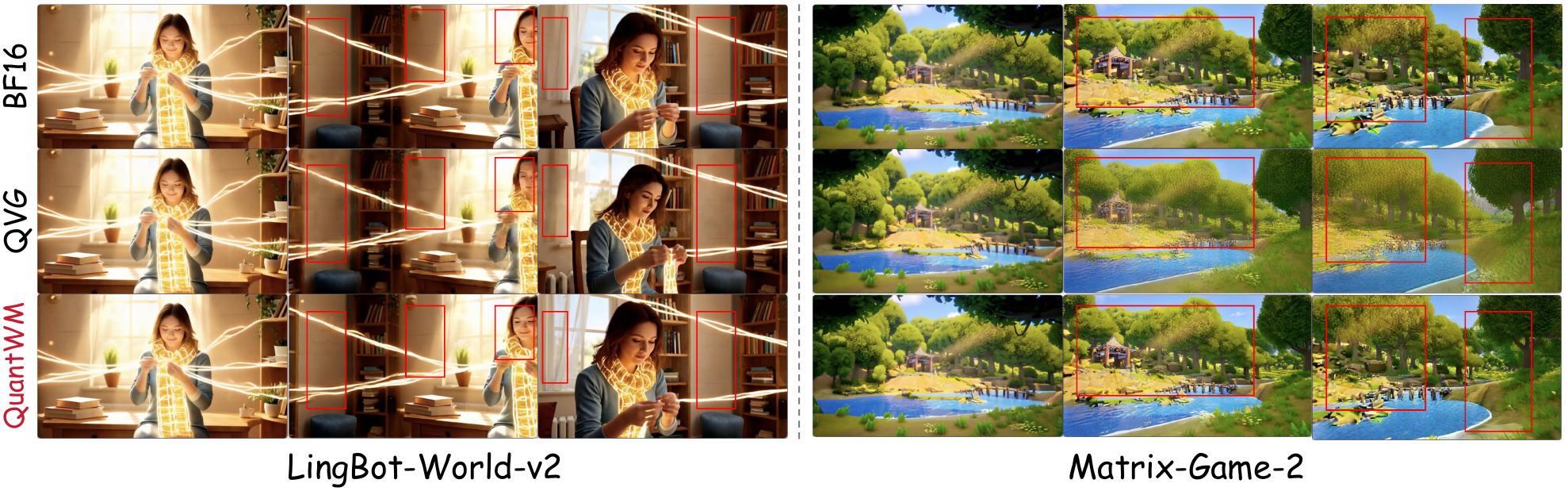}%
\caption{Visual quality comparison of \textbf{Quant\textit{WM}}, QVG and BF16. QVG suffers from temporal flickering and visual degradation, while our \textbf{Quant\textit{WM}} preserves clearer details and visual quality. 
}
\label{videos}
\end{figure*}

\begin{table*}[t]
\centering
\caption{
Visual quality and VBench comparison results of \textbf{Quant\textit{WM}} and baselines on 480p videos.
\textbf{Quant\textit{WM}} significantly improves visual quality while maintaining strong VBench performance.}
\label{480p_main_results}

\setlength{\tabcolsep}{3.2pt}
\renewcommand{\arraystretch}{1.08}
\scriptsize
\resizebox{\linewidth}{!}{
\begin{tabular}{c|c|ccc|cccccc}
\toprule

\multirow{2}{*}{\textbf{Model}} &
\multirow{2}{*}{\textbf{Method}} &
\multicolumn{3}{c|}{\cellcolor{blue!7}\textbf{Frame-level Quality}} &
\multicolumn{6}{c}{\cellcolor{blue!7}\textit{VBench}} \\

\cmidrule(lr){3-5}
\cmidrule(lr){6-11}

& &
\cellcolor{blue!7}\textbf{PSNR} $\uparrow$ &
\cellcolor{blue!7}\textbf{SSIM} $\uparrow$ &
\cellcolor{blue!7}\textbf{LPIPS} $\downarrow$ &
\cellcolor{blue!7}\textit{Subject} $\uparrow$ &
\cellcolor{blue!7}\textit{Background} $\uparrow$ &
\cellcolor{blue!7}\textit{Aesthetic} $\uparrow$ &
\cellcolor{blue!7}\textit{Image} $\uparrow$ &
\cellcolor{blue!7}\textit{Temporal} $\uparrow$ &
\cellcolor{blue!7}\textbf{\textit{Rank}} $\downarrow$ \\

\midrule

\multirow{4}{*}{Matrix-Game-2}
& BF16
& $\infty$ & 1.0000 & 0.0000
& 0.9104 & 0.9423 & 0.5846 & 0.6952 & 0.9553
& -- \\

& KIVI
& 19.298 & 0.6557 & 0.1797
& 0.9097 & 0.9418 & 0.5841 & 0.6914 & \textbf{0.9603}
& 1.80 \\

& QVG
& 17.749 & 0.5651 & 0.2851
& 0.8894 & 0.9320 & 0.5831 & 0.6603 & 0.9558
& 3.00 \\

& \ourcell{\textbf{Ours}}
& \ourcell{\textbf{20.615}}
& \ourcell{\textbf{0.7212}}
& \ourcell{\textbf{0.1443}}
& \ourcell{\textbf{0.9109}}
& \ourcell{\textbf{0.9421}}
& \ourcell{\textbf{0.5883}}
& \ourcell{\textbf{0.6933}}
& \ourcell{0.9561}
& \ourcell{\textbf{1.20}} \\

\midrule

\multirow{4}{*}{LingBot-v2}
& BF16
& $\infty$ & 1.0000 & 0.0000
& 0.8877 & 0.9126 & 0.5949 & 0.6962 & 0.9314
& -- \\

& KIVI
& 14.799 & 0.5950 & 0.2508
& 0.8869 & \textbf{0.9137} & 0.5883 & 0.6892 & 0.9320
& 2.40 \\

& QVG
& 14.414 & 0.5726 & 0.2779
& 0.8860 & 0.9127 & 0.5912 & 0.6929 & \textbf{0.9329}
& 2.20 \\

& \ourcell{\textbf{Ours}}
& \ourcell{\textbf{16.138}}
& \ourcell{\textbf{0.6688}}
& \ourcell{\textbf{0.1893}}
& \ourcell{\textbf{0.8894}}
& \ourcell{0.9134}
& \ourcell{\textbf{0.5947}}
& \ourcell{\textbf{0.6953}}
& \ourcell{0.9323}
& \ourcell{\textbf{1.40}} \\

\midrule

\multirow{4}{*}{HY-World 1.5}
& BF16
& $\infty$ & 1.0000 & 0.0000
& 0.9588 & 0.9520 & 0.6378 & 0.7209 & 0.9562
& -- \\

& KIVI
& 17.677 & 0.7165 & 0.1540
& 0.9520 & 0.9494 & 0.6362 & 0.7049 & \textbf{0.9702}
& 2.20 \\

& QVG
& 16.118 & 0.6559 & 0.2114
& 0.9503 & 0.9488 & \textbf{0.6394} & 0.7122 & 0.9585
& 2.20 \\

& \ourcell{\textbf{Ours}}
& \ourcell{\textbf{18.449}}
& \ourcell{\textbf{0.7661}}
& \ourcell{\textbf{0.1082}}
& \ourcell{\textbf{0.9600}}
& \ourcell{\textbf{0.9528}}
& \ourcell{0.6366}
& \ourcell{\textbf{0.7196}}
& \ourcell{0.9584}
& \ourcell{\textbf{1.60}} \\

\midrule

\multirow{4}{*}{LongCat-Video}
& BF16
& $\infty$ & 1.0000 & 0.0000
& 0.9747 & 0.9605 & 0.6148 & 0.6993 & 0.9461
& -- \\

& KIVI
& 24.448 & 0.9180 & 0.0399
& \textbf{0.9754} & 0.9610 & 0.6141 & 0.6695 & 0.9475
& 2.20 \\

& QVG
& 20.683 & 0.8556 & 0.0784
& 0.9739 & 0.9603 & 0.6144 & 0.7001 & 0.9463
& 2.60 \\

& \ourcell{\textbf{Ours}}
& \ourcell{\textbf{25.508}}
& \ourcell{\textbf{0.9286}}
& \ourcell{\textbf{0.0332}}
& \ourcell{0.9748}
& \ourcell{\textbf{0.9615}}
& \ourcell{\textbf{0.6154}}
& \ourcell{\textbf{0.7002}}
& \ourcell{\textbf{0.9489}}
& \ourcell{\textbf{1.20}} \\

\midrule

\multirow{4}{*}{Causal-Forcing}
& BF16
& $\infty$ & 1.0000 & 0.0000
& 0.9327 & 0.9421 & 0.6348 & 0.7028 & 0.9534
& -- \\

& KIVI
& 15.585 & 0.7201 & 0.1775
& 0.9334 & 0.9426 & 0.6343 & 0.7034 & 0.9540
& 2.20 \\

& QVG
& 14.887 & 0.6773 & 0.2185
& 0.9311 & 0.9412 & 0.6317 & 0.6993 & 0.9546
& 2.80 \\

& \ourcell{\textbf{Ours}}
& \ourcell{\textbf{16.283}}
& \ourcell{\textbf{0.7464}}
& \ourcell{\textbf{0.1554}}
& \ourcell{\textbf{0.9342}}
& \ourcell{\textbf{0.9427}}
& \ourcell{\textbf{0.6354}}
& \ourcell{\textbf{0.7044}}
& \ourcell{\textbf{0.9558}}
& \ourcell{\textbf{1.00}} \\

\bottomrule
\end{tabular}
}
\end{table*}

\begin{table*}[t]
\centering
\caption{
Visual quality and VBench comparison results of \textbf{Quant\textit{WM}} and baselines on 720p videos.}
\label{720p_results}

\setlength{\tabcolsep}{3.2pt}
\renewcommand{\arraystretch}{1.08}
\scriptsize

\resizebox{\linewidth}{!}{
\begin{tabular}{c|c|ccc|cccccc}
\toprule

\multirow{2}{*}{\textbf{Model}} &
\multirow{2}{*}{\textbf{Method}} &
\multicolumn{3}{c|}{\cellcolor{blue!7}\textbf{Frame-level Quality}} &
\multicolumn{6}{c}{\cellcolor{blue!7}\textit{VBench}} \\

\cmidrule(lr){3-5}
\cmidrule(lr){6-11}

& &
\cellcolor{blue!7}\textbf{PSNR} $\uparrow$ &
\cellcolor{blue!7}\textbf{SSIM} $\uparrow$ &
\cellcolor{blue!7}\textbf{LPIPS} $\downarrow$ &
\cellcolor{blue!7}\textit{Subject} $\uparrow$ &
\cellcolor{blue!7}\textit{Background} $\uparrow$ &
\cellcolor{blue!7}\textit{Aesthetic} $\uparrow$ &
\cellcolor{blue!7}\textit{Image} $\uparrow$ &
\cellcolor{blue!7}\textit{Temporal} $\uparrow$ &
\cellcolor{blue!7}\textbf{\textit{Rank}} $\downarrow$ \\

\midrule

\multirow{4}{*}{Matrix-Game-2}
& BF16
& $\infty$ & 1.0000 & 0.0000
& 0.9096 & 0.9490 & 0.5875 & 0.7038 & 0.9550
& -- \\

& KIVI
& 19.200 & 0.6659 & 0.2048
& 0.9081 & 0.9479 & 0.5888 & 0.7003 & \textbf{0.9601}
& 1.80 \\

& QVG
& 17.721 & 0.5871 & 0.3047
& 0.8884 & 0.9373 & 0.5840 & 0.6679 & 0.9556
& 3.00 \\

& \ourcell{\textbf{Ours}}
& \ourcell{\textbf{20.568}}
& \ourcell{\textbf{0.7234}}
& \ourcell{\textbf{0.1651}}
& \ourcell{\textbf{0.9100}}
& \ourcell{\textbf{0.9488}}
& \ourcell{\textbf{0.5914}}
& \ourcell{\textbf{0.7020}}
& \ourcell{0.9559}
& \ourcell{\textbf{1.20}} \\

\midrule

\multirow{4}{*}{LingBot-v2}
& BF16
& $\infty$ & 1.0000 & 0.0000
& 0.8984 & 0.9202 & 0.6224 & 0.7211 & 0.9362
& -- \\

& KIVI
& 15.134 & 0.6317 & 0.2520
& 0.8965 & \textbf{0.9224} & 0.6180 & 0.7165 & 0.9364
& 2.40 \\

& QVG
& 13.927 & 0.5958 & 0.3220
& 0.8906 & 0.9195 & 0.6181 & 0.7175 & \textbf{0.9367}
& 2.20 \\

& \ourcell{\textbf{Ours}}
& \ourcell{\textbf{15.564}}
& \ourcell{\textbf{0.6540}}
& \ourcell{\textbf{0.2267}}
& \ourcell{\textbf{0.8994}}
& \ourcell{0.9213}
& \ourcell{\textbf{0.6214}}
& \ourcell{\textbf{0.7207}}
& \ourcell{0.9365}
& \ourcell{\textbf{1.40}} \\

\midrule

\multirow{4}{*}{HY-World 1.5}
& BF16
& $\infty$ & 1.0000 & 0.0000
& 0.9592 & 0.9629 & 0.6556 & 0.7236 & 0.9659
& -- \\

& KIVI
& 17.643 & 0.7338 & 0.1747
& 0.9523 & 0.9593 & 0.6518 & 0.7119 & \textbf{0.9699}
& 2.20 \\

& QVG
& 16.085 & 0.6810 & 0.2358
& 0.9505 & 0.9583 & \textbf{0.6575} & 0.7181 & 0.9678
& 2.20 \\

& \ourcell{\textbf{Ours}}
& \ourcell{\textbf{18.423}}
& \ourcell{\textbf{0.7779}}
& \ourcell{\textbf{0.1257}}
& \ourcell{\textbf{0.9602}}
& \ourcell{\textbf{0.9636}}
& \ourcell{0.6551}
& \ourcell{\textbf{0.7229}}
& \ourcell{0.9667}
& \ourcell{\textbf{1.60}} \\

\midrule

\multirow{4}{*}{LongCat-Video}
& BF16
& $\infty$ & 1.0000 & 0.0000
& 0.9717 & 0.9678 & 0.6327 & 0.6975 & 0.9609
& -- \\

& KIVI
& 22.896 & 0.8805 & 0.0708
& \textbf{0.9719} & \textbf{0.9675} & 0.6334 & 0.6962 & 0.9610
& 1.90 \\

& QVG
& 20.988 & 0.8246 & 0.1098
& 0.9714 & 0.9671 & 0.6325 & 0.6963 & 0.9609
& 2.80 \\

& \ourcell{\textbf{Ours}}
& \ourcell{\textbf{25.127}}
& \ourcell{\textbf{0.9094}}
& \ourcell{\textbf{0.0503}}
& \ourcell{0.9715}
& \ourcell{\textbf{0.9675}}
& \ourcell{\textbf{0.6340}}
& \ourcell{\textbf{0.6978}}
& \ourcell{\textbf{0.9620}}
& \ourcell{\textbf{1.30}} \\

\midrule

\multirow{4}{*}{Causal-Forcing}
& BF16
& $\infty$ & 1.0000 & 0.0000
& 0.9345 & 0.9514 & 0.6340 & 0.7038 & 0.9525
& -- \\

& KIVI
& 15.394 & 0.7349 & 0.2013
& 0.9335 & 0.9517 & 0.6349 & 0.7023 & 0.9532
& 2.20 \\

& QVG
& 14.771 & 0.6996 & 0.2390
& 0.9321 & 0.9498 & 0.6321 & 0.6982 & 0.9543
& 2.80 \\

& \ourcell{\textbf{Ours}}
& \ourcell{\textbf{16.231}}
& \ourcell{\textbf{0.7628}}
& \ourcell{\textbf{0.1716}}
& \ourcell{\textbf{0.9354}}
& \ourcell{\textbf{0.9521}}
& \ourcell{\textbf{0.6352}}
& \ourcell{\textbf{0.7034}}
& \ourcell{\textbf{0.9554}}
& \ourcell{\textbf{1.00}} \\

\bottomrule
\end{tabular}
}
\end{table*}

\section{Experiments}

\subsection{Experimental Setup}
\paragraph{Models}
To validate the effectiveness of \textbf{Quant\textit{WM}}, we conduct extensive experiments on 3 open-sourced video world models, including Matrix-Game-2 \citep{he2025matrix}, HY-World 1.5 \citep{sun2025worldplay} and LingBot-World-v2 \citep{gao2026infinite}. We also include 2 autoregressive video generation models LongCat-Video \citep{team2025longcat} and Causal-Forcing \citep{zhu2026causal} to evaluate the generalization. We conduct the main evaluation on $480$p and $720$p generated videos with 93 frames. 

\paragraph{Evaluations}
We select Quant VideoGen \citep{xi2026quant}, the most relevant 2-bit KV cache quantization method for video generation, as the main baseline. The representative LLMs quantization method KIVI \citep{liu2024kivi} is also included into comparison. We mainly evaluate the visual degradation introduced by quantization using frame-level visual quality and qualitative visualizations. During experiments, we compute them over all frames of each video. We also report performance on VBench \citep{huang2024vbench}. Although it may not fully capture the visual degradation studied in this work, it is still an important benchmark for evaluating the overall quality of generated videos. Following QVG, we use the prompt suite from MovieGen Benchmark \citep{polyak2024movie}.\footnote{\texttt{facebookresearch/MovieGenBench}} For Matrix-Game-2, we use the official first frames combined with 7--8 different action sequences as inputs.\footnote{\texttt{SkyworkAI/Matrix-Game/tree/main/Matrix-Game-2/demo\_images}}. For fair comparison, we reproduce all the baselines with their official repository, follow the native KV cache convention of each model and apply all methods to the same cached tensors.

\paragraph{Implementation}
We implement \textbf{Quant\textit{WM}} in PyTorch and conduct all experiments on NVIDIA A800 80GB GPUs. Following QVG, we apply streaming chunk-wise KV cache quantization. We use $256$ centroids and quantization groupsize $64$ (QVG with 64 and KIVI with 32), with BF16 centroids and INT2 quantization for the residuals. For PSAC, it extracts the top-8 eigenvectors $\mathbf{U}_r$ of $\mathbf{M}_{h}^{t-1}$ as the principal Query subspace. We store $\mathbf{U}_r$ in BF16 and quantize $\mathbf{C}=\mathbf{E}\mathbf{U}_r$ to INT8.

\subsection{Main Results and Visualizations}

\paragraph{Visual quality evaluation}
As shown in Table \ref{480p_main_results}, \ref{720p_results} and Figure \ref{videos}, 
\textbf{Quant\textit{WM}} consistently improves frame-level visual quality over QVG across all five models while visibly reducing temporal flickering, blurring and artifacts. 
For example, on LongCat-Video, PSNR improves from $20.683$ to $25.508$ and LPIPS decreases from $0.0784$ to $0.0332$. 
Although KIVI benefits from a finer groupsize and a BF16 recent-window cache, \textbf{Quant\textit{WM}} still achieves the best results across all models. 
Long video generation results and more visualizations are provided in Appendix \ref{more-results}.

\paragraph{Video benchmark evaluation}
\textbf{Quant\textit{WM}} also maintains strong performance and the best average rank on VBench across all five models. Notably, the Temporal scores are very close among different methods, even when QVG exhibits clear temporal flickering in Figure \ref{videos}, indicating that this metric does not fully capture such visual degradation. Considering both metrics, \textbf{Quant\textit{WM}} demonstrates stronger ability to preserve visual quality under 2-bit KV cache quantization on video world models.

\begin{table}[t]
\centering
\caption{
Top-1 temporal-spatial token selection shift ratio on 480p videos.
\textbf{Quant\textit{WM}} significantly reduces the token-selection shifts caused by 2-bit Key quantization.
}
\label{top1_shift}

\setlength{\tabcolsep}{5.0pt}
\renewcommand{\arraystretch}{1.08}
\small

\begin{tabular}{c|ccccc}
\toprule
\cellcolor{blue!7}\textbf{Method} &
\cellcolor{blue!7}\textbf{Matrix-Game-2} &
\cellcolor{blue!7}\textbf{LingBot-v2} &
\cellcolor{blue!7}\textbf{HY-World 1.5} &
\cellcolor{blue!7}\textbf{LongCat-Video} &
\cellcolor{blue!7}\textbf{Causal-Forcing} \\
\midrule

BF16
& 0.0000
& 0.0000
& 0.0000
& 0.0000
& 0.0000 \\

QVG
& 0.5388
& 0.5491
& 0.6136
& 0.5034
& 0.3728 \\

\ourcell{\textbf{Ours}}
& \ourcell{\textbf{0.1019}}
& \ourcell{\textbf{0.1600}}
& \ourcell{\textbf{0.1392}}
& \ourcell{\textbf{0.2225}}
& \ourcell{\textbf{0.1420}} \\

\bottomrule
\end{tabular}
\end{table}

\begin{figure*}[t]
\centering
\includegraphics[width=4.6in]{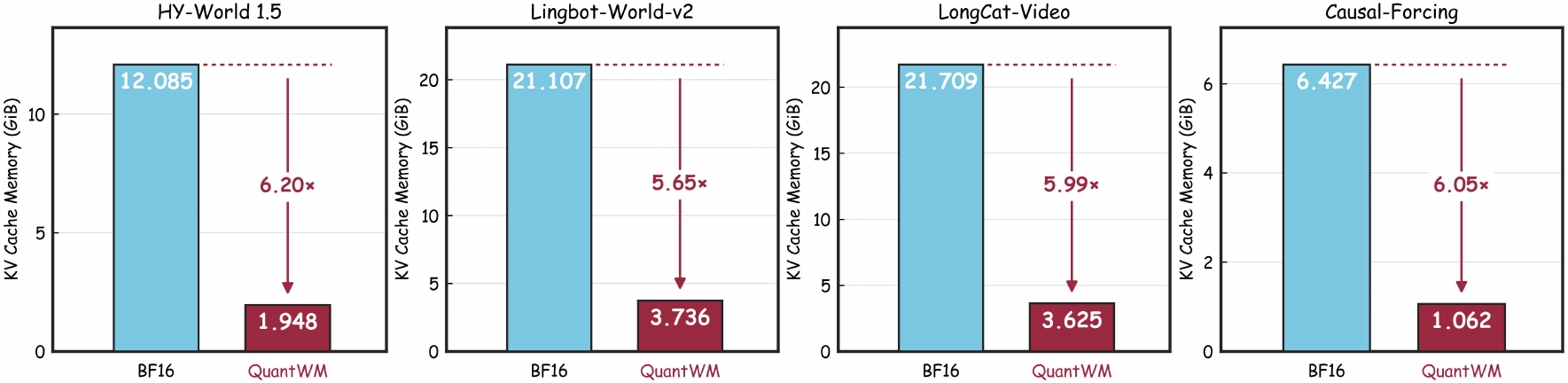}%
\caption{KV cache memory comparison of BF16 and \textbf{Quant\textit{WM}} for 93-frame video generation. \textbf{Quant\textit{WM}} achieves up to $6.20\times$ KV cache memory compression across different models.
}
\label{memory}
\end{figure*}

\subsection{Temporal-spatial token selection shifts ratio}
To further verify whether \textbf{Quant\textit{WM}} preserves attention logits under 2-bit quantization, we evaluate the top-1 temporal-spatial token selection shift ratio compared with BF16. As shown in Table \ref{top1_shift}, QVG causes frequent token-selection changes across all five models, with shift ratios up to $61.36\%$. In contrast, reduces the average shift ratio from $51.55\%$ to $15.31\%$. These results further proves that our \textbf{Quant\textit{WM}} successfully reduces the quantization perturbation to the attention.

\subsection{Ablation Study}

\begin{wraptable}{r}{8.4cm}
\vspace{-1.3em}
\centering
\caption{
Ablation study of \textbf{Quant\textit{WM}} on HY-World 1.5.
}
\label{ablation}
\setlength{\tabcolsep}{4pt}
\renewcommand{\arraystretch}{1.08}
\small

\begin{tabular}{cc|cccc}
\toprule

\cellcolor{blue!7}\textbf{QSAC} &
\cellcolor{blue!7}\textbf{PSAC} &
\cellcolor{blue!7}\textbf{PSNR} $\uparrow$ &
\cellcolor{blue!7}\textbf{SSIM} $\uparrow$ &
\cellcolor{blue!7}\textbf{LPIPS} $\downarrow$ &
\cellcolor{blue!7}\textbf{Token Shift} $\downarrow$ \\

\midrule

$\times$
& $\times$
& 16.118
& 0.6559
& 0.2114
& 0.6136 \\

$\checkmark$
& $\times$
& 17.990
& 0.7423
& 0.1285
& 0.3817 \\

$\times$
& $\checkmark$
& 17.849
& 0.7358
& 0.1312
& 0.2290 \\

\ourcell{$\checkmark$}
& \ourcell{$\checkmark$}
& \ourcell{\textbf{18.449}}
& \ourcell{\textbf{0.7661}}
& \ourcell{\textbf{0.1082}}
& \ourcell{\textbf{0.1392}} \\

\bottomrule
\end{tabular}
\end{wraptable}

To demonstrate the contributions of QSAC and PSAC, We conduct an ablation study on HY-World 1.5. As shown in Table \ref{ablation}, QSAC consistently improves frame-level quality metrics while reducing the token shift ratio from $61.36\%$ to $38.17\%$. This demonstrates that selecting centroids according to Query sensitivity and residual quantization difficulty effectively reduces harmful Key quantization perturbations. PSAC further shows a stronger effect on attention preservation, where it reduces the token Shift to $22.9\%$ even without QSAC. Finally, combining QSAC and PSAC achieves the best overall performance. 

\subsection{Inference Efficiency}

\paragraph{KV cache memory usage}
We first evaluate the KV cache memory consumption during 93-frame video generation. As shown in Figure \ref{memory}, \textbf{Quant\textit{WM}} consistently reduces the KV cache footprint across different models, achieving up to $6.20\times$ compression. For example, on LongCat-Video, \textbf{Quant\textit{WM}} reduces the KV cache memory from $21.709$ GiB to $3.625$ GiB. These results demonstrate the effectiveness of \textbf{Quant\textit{WM}} in reducing the memory cost during generation.

\paragraph{End-to-end latency}

\begin{wraptable}{r}{6.6cm}
\vspace{-1.4em}
\centering
\caption{
Inference latency comparison between BF16 and \textbf{Quant\textit{WM}}.
}
\label{latency}

\setlength{\tabcolsep}{5.5pt}
\renewcommand{\arraystretch}{1.08}
\small

\begin{tabular}{c|cc}
\toprule
\cellcolor{blue!7}\textbf{Model} &
\cellcolor{blue!7}\textbf{BF16 (s)} &
\cellcolor{blue!7}\textbf{\textbf{Quant\textit{WM}} (s)} \\
\midrule

LingBot-World-v2
& 193.56
& 204.09 {\color{gray}\scriptsize 5.44\%$\uparrow$} \\

HY-World 1.5
& 112.17
& 118.75 {\color{gray}\scriptsize 5.87\%$\uparrow$} \\

LongCat-Video
& 255.35
& 209.94 {\color{gray}\scriptsize \textbf{17.78}\%$\downarrow$} \\

\bottomrule
\end{tabular}
\end{wraptable}

As shown in Table \ref{latency}, \textbf{Quant\textit{WM}} introduces only limited latency overhead across different models. The generation latency increases by only $5.44\%$ on LingBot-World-v2 and $5.87\%$ on HY-World 1.5, while it is reduced by $17.78\%$ on LongCat-Video. These results indicate that \textbf{Quant\textit{WM}} achieves significant memory compression with limited impact on end-to-end generation latency. Detailed implementations and analysis of real-quant inference system design are provided in Appendix \ref{system}.

\section{Conclusion}
In this paper, we investigate the overlooked visual degradation caused by KV cache quantization in video world models. We show that Key quantization is more sensitive than Value because Key perturbations can influence attention logits and shift temporal-spatial token selection. Based on this observation, we propose \textbf{Quant\textit{WM}}, a training-free 2-bit KV cache quantization framework that considers attention preservation. \textbf{Quant\textit{WM}} includes quantization-sensitivity-aware clustering (QSAC), which incorporates Query sensitivity and residual quantization difficulty into centroid selection, and principal-subspace attention compensation (PSAC), which directly corrects the remaining logit perturbations along dominant Query directions. Experiments across multiple video world models demonstrate that \textbf{Quant\textit{WM}} effectively alleviates temporal flickering and visual degradation, while achieving high KV cache memory compression with limited overheads.

\bibliography{iclr2027_conference}
\bibliographystyle{iclr2027_conference}

\clearpage

\appendix

\section{Related Works}

\subsection{Video Generation and World Models}
Diffusion-based video generation has advanced rapidly through large-scale spatial-temporal Transformers, which significantly improves visual fidelity and temporal consistency \citep{singer2022make,zheng2024open,yang2025cogvideox}. Recent autoregressive approaches further enable long-form and streaming generation by producing videos chunk by chunk under causal attention \citep{huang2026self,liu2026rolling,zhu2026causal,yang2025longlive}. Building on these advances, video world models incorporate action and camera controls to support real-time interaction with generated environments \citep{alonso2024diffusion,he2025matrix,gao2026infinite,valevski2025diffusion}. Long-term consistency has also motivated explicit memory mechanisms that retrieve or reconstruct relevant historical observations \citep{xiao2026worldmem,sun2025worldplay,wang2026matrix}. Nevertheless, causal video models commonly storage historical representations through KV caches. Since each generated chunk introduces numerous spatial-temporal tokens, the cache grows rapidly with the cached context and will finally dominate GPU memory, making KV cache efficiency a central bottleneck for video generation and world modeling.

\subsection{Model Quantization}
Quantization \citep{nagel2020up,liu2024llm} reduces storage and inference overheads by transfer high-precision tensors to low-bit values. Post-training quantization \citep{shao2024omniquant,huang2024slim,zhao2025lrquant} has become particularly attractive because it avoids costly model retraining. Representative methods include GPTQ \citep{frantar2022gptq}, which uses approximate second-order information for weight reconstruction, SmoothQuant \citep{xiao2023smoothquant}, which migrates activation outliers into weights for W8A8 quantization, and AWQ \citep{lin2024awq}, which protects activation-salient weight channels. These methods primarily focus on efficient inference by compressing model weights and activations.

KV cache quantization further reduces the memory accumulated during autoregressive inference. KIVI \citep{liu2024kivi} applies asymmetric 2-bit quantization to Keys and Values, while KVQuant \citep{hooper2024kvquant} introduces pre-RoPE Key quantization, non-uniform datatypes, and outlier isolation. CommVQ \citep{li2025commvq} instead compresses KV caches using RoPE-compatible learned codebooks. These methods mainly target language models. More recently, Quant-VideoGen (QVG) \citep{xi2026quant} extends 2-bit KV cache quantization to autoregressive video diffusion through semantic-aware smoothing and residual quantization, which achieves nearly lossless performance on video benchmarks. However, we find that such benchmark-level results ignores severe temporal flickering and visual artifacts. Recent work \citep{tuncer2026quantized} discovers a systematic softmax attention bias introduced by quantized Keys and corrects it at the attention-score level. In contrast, we focus on quantization perturbations of $QK^\top$ and temporal-spatial token selection, and address them through Query-sensitive centroid assignment and principal-subspace error compensation to improve temporal consistency.

\begin{table}[t]
\centering
\caption{
VBench results on 1-minute video generation.
\textbf{Quant\textit{WM}} consistently improves video quality over QVG under long-video generation.
}
\label{long_video}

\setlength{\tabcolsep}{3.2pt}
\renewcommand{\arraystretch}{1.08}
\small

\begin{tabular}{c|c|cccccc}
\toprule

\multirow{2}{*}{\textbf{Model}} &
\multirow{2}{*}{\textbf{Method}} &
\multicolumn{6}{c}{\cellcolor{blue!7}\textit{VBench}} \\

\cmidrule(lr){3-8}

& &
\cellcolor{blue!7}\textit{Subject} $\uparrow$ &
\cellcolor{blue!7}\textit{Background} $\uparrow$ &
\cellcolor{blue!7}\textit{Aesthetic} $\uparrow$ &
\cellcolor{blue!7}\textit{Image} $\uparrow$ &
\cellcolor{blue!7}\textit{Temporal} $\uparrow$ &
\cellcolor{blue!7}\textbf{\textit{Rank}} $\downarrow$ \\

\midrule

\multirow{3}{*}{Matrix-Game-2}
& BF16
& 0.7287 & 0.8659 & 0.4475 & 0.6767 & 0.9857
& -- \\

& QVG
& 0.7280 & 0.8675 & 0.4439 & 0.5275 & 0.9858
& 2.00 \\

& \ourcell{\textbf{Ours}}
& \ourcell{\textbf{0.7328}}
& \ourcell{\textbf{0.8693}}
& \ourcell{\textbf{0.4523}}
& \ourcell{\textbf{0.6639}}
& \ourcell{\textbf{0.9878}}
& \ourcell{\textbf{1.00}} \\

\midrule

\multirow{3}{*}{LingBot-v2}
& BF16
& 0.9063 & 0.9318 & 0.6614 & 0.7593 & 0.9786
& -- \\

& QVG
& 0.8946 & 0.9280 & 0.6404 & 0.7620 & \textbf{0.9774}
& 1.80 \\

& \ourcell{\textbf{Ours}}
& \ourcell{\textbf{0.9121}}
& \ourcell{\textbf{0.9353}}
& \ourcell{\textbf{0.6567}}
& \ourcell{\textbf{0.7633}}
& \ourcell{0.9767}
& \ourcell{\textbf{1.20}} \\

\midrule

\multirow{3}{*}{HY-World 1.5}
& BF16
& 0.7051 & 0.8242 & 0.4730 & 0.6247 & 0.9656
& -- \\

& QVG
& 0.6591 & 0.8025 & 0.4423 & 0.4756 & \textbf{0.9795}
& 1.80 \\

& \ourcell{\textbf{Ours}}
& \ourcell{\textbf{0.7100}}
& \ourcell{\textbf{0.8269}}
& \ourcell{\textbf{0.4895}}
& \ourcell{\textbf{0.6242}}
& \ourcell{0.9715}
& \ourcell{\textbf{1.20}} \\

\midrule

\multirow{3}{*}{LongCat-Video}
& BF16
& 0.8340 & 0.8939 & 0.6334 & 0.7396 & 0.9593
& -- \\

& QVG
& 0.8253 & 0.8920 & 0.6222 & 0.7323 & 0.9562
& 2.00 \\

& \ourcell{\textbf{Ours}}
& \ourcell{\textbf{0.8378}}
& \ourcell{\textbf{0.8957}}
& \ourcell{\textbf{0.6443}}
& \ourcell{\textbf{0.7359}}
& \ourcell{\textbf{0.9592}}
& \ourcell{\textbf{1.00}} \\

\midrule

\multirow{3}{*}{Causal-Forcing}
& BF16
& 0.6903 & 0.8445 & 0.4596 & 0.5647 & 0.9605
& -- \\

& QVG
& 0.7052 & \textbf{0.8213} & 0.5032 & 0.6005 & 0.9497
& 1.80 \\

& \ourcell{\textbf{Ours}}
& \ourcell{\textbf{0.7184}}
& \ourcell{0.8155}
& \ourcell{\textbf{0.5070}}
& \ourcell{\textbf{0.6566}}
& \ourcell{\textbf{0.9614}}
& \ourcell{\textbf{1.20}} \\

\bottomrule
\end{tabular}
\end{table}

\section{More Evaluation Results}
\label{more-results}


\subsection{Results on Long-video Generation}

We further evaluate \textbf{Quant\textit{WM}} under 1-minute long-horizon video generation. Long-form autoregressive generation is particularly challenging for low-bit KV cache quantization, since quantization perturbations can accumulate over successive generation steps and lead to larger deviations from the BF16 trajectory. Therefore, we focus on the overall video quality measured by VBench in this setting. As shown in Table \ref{long_video}, \textbf{Quant\textit{WM}} outperforms QVG on most VBench dimensions and achieves the best average rank across all five models. For example, on HY-World 1.5, the Image score improves from $0.4756$ to $0.6242$. Overall, \textbf{Quant\textit{WM}} maintains strong video quality under longer generation, demonstrating better robustness to accumulated quantization errors.

\begin{figure*}[t]
\centering
\subfloat[LingBot-World-v2]{\includegraphics[width=5.5in]{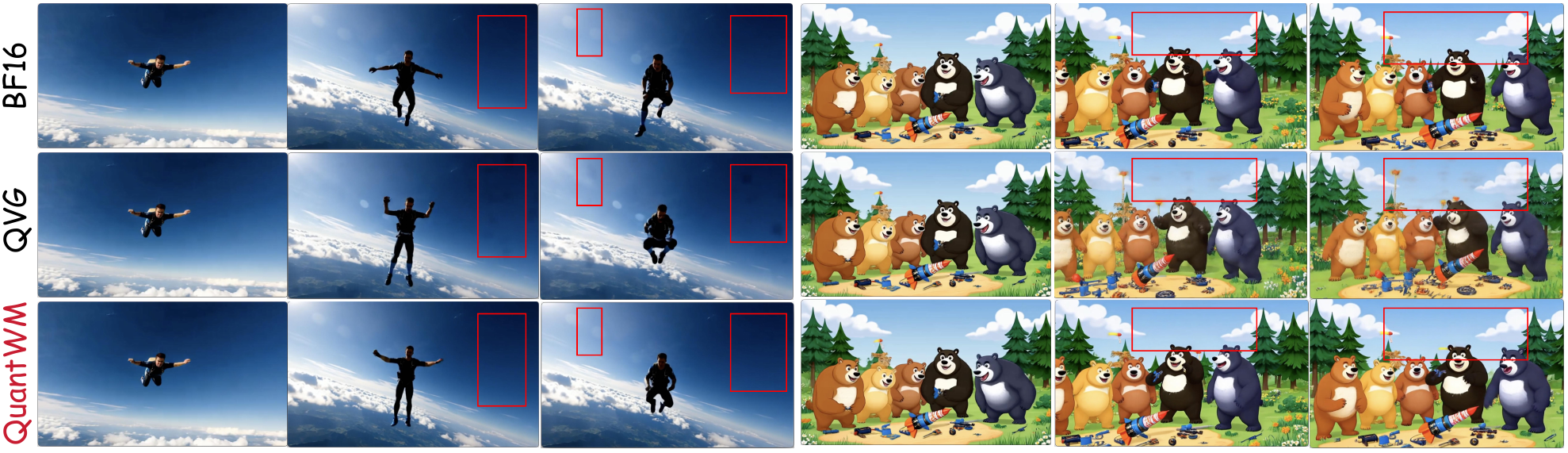}}%
\\
\subfloat[HY-World 1.5]{\includegraphics[width=5.5in]{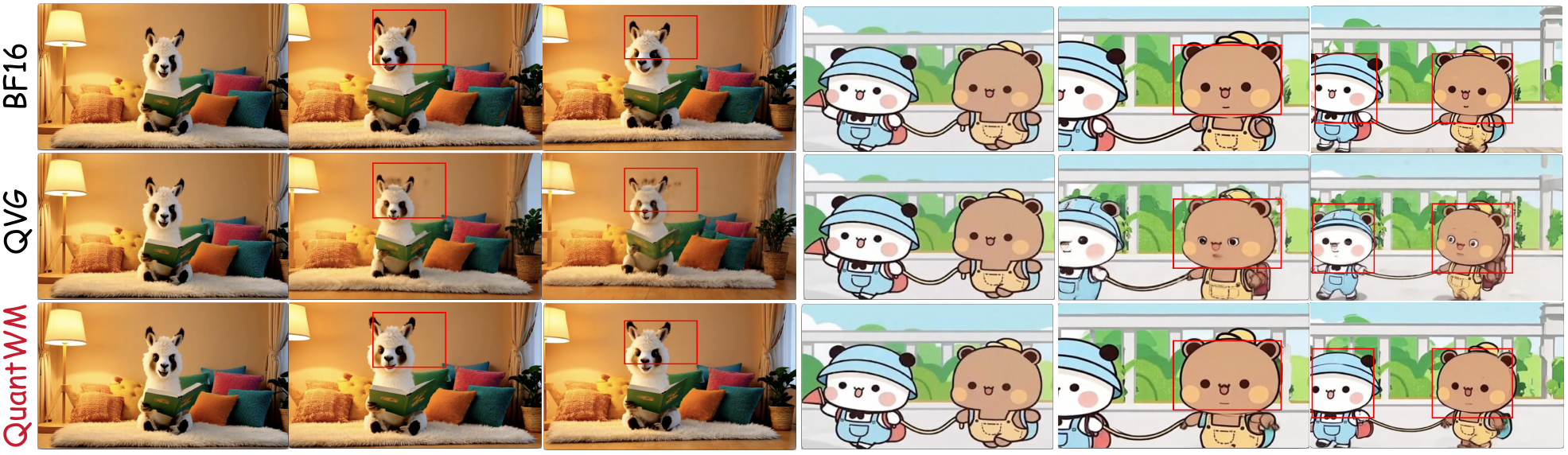}}
\\
\subfloat[Matrix-Game-2]{\includegraphics[width=5.5in]{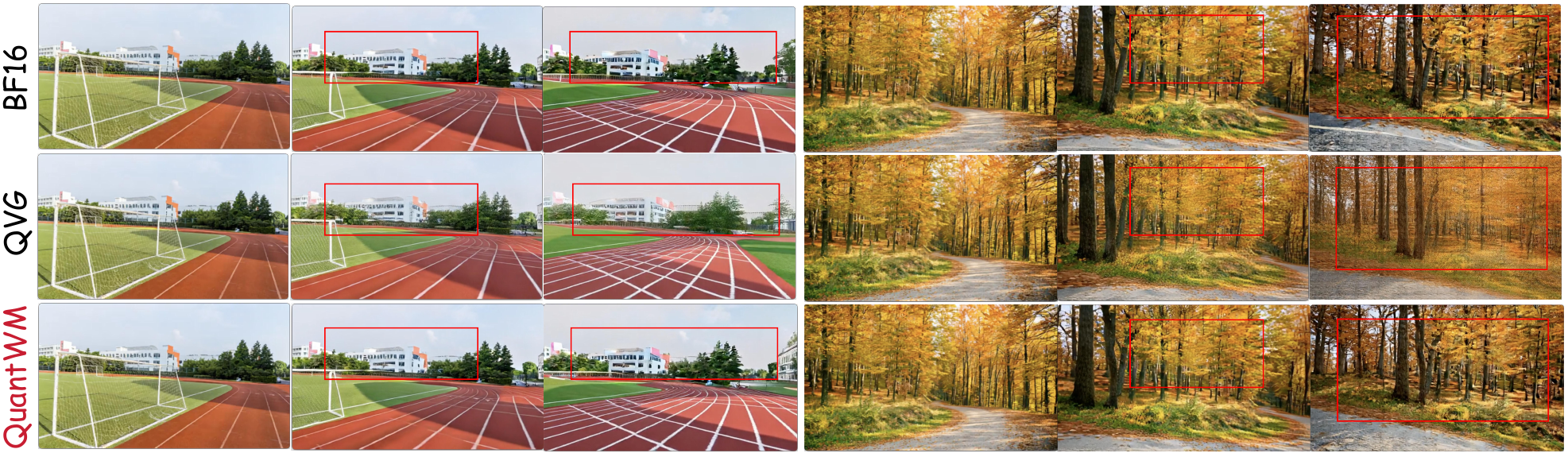}}%
\\
\subfloat[Causal-Forcing]{\includegraphics[width=5.5in]{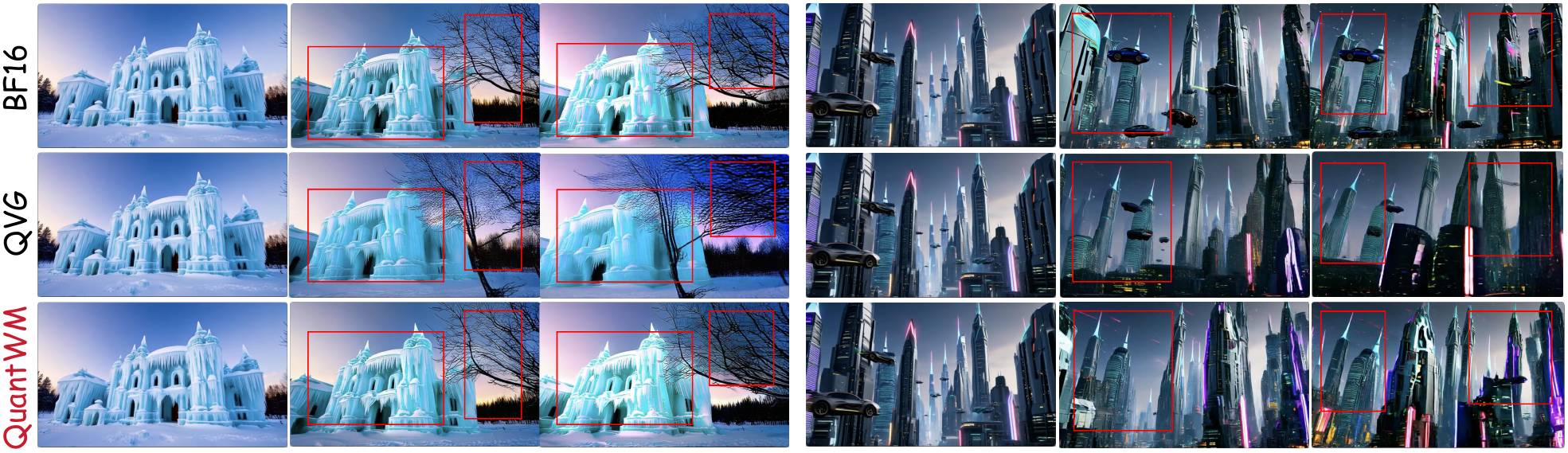}}%
\\
\caption{More visual comparisons of BF16, QVG and \textbf{Quant\textit{WM}} across different video generation and world models. Please refer to the project page for detailed video comparison.}
\label{more_visualizations}
\end{figure*}

\subsection{More Visualizations of Generated Videos}

We provide additional qualitative comparisons across different video generation and world models. As shown in Figure \ref{more_visualizations}, QVG frequently introduces temporal flickering, blurring, and local visual artifacts after 2-bit KV cache quantization. In contrast, \textbf{Quant\textit{WM}} better preserves object appearance, spatial details, and temporal consistency, which produces results that remain closer to BF16. These examples further confirm that the temporal consistency improvements of \textbf{Quant\textit{WM}} are consistent across different models and generation scenarios.

\section{Analysis and Illustrations on QSAC and PSAC}

\subsection{Query Energy Distribution on More Models}
\label{app:query_energy}

\begin{figure*}[t]
\centering
\includegraphics[width=5in]{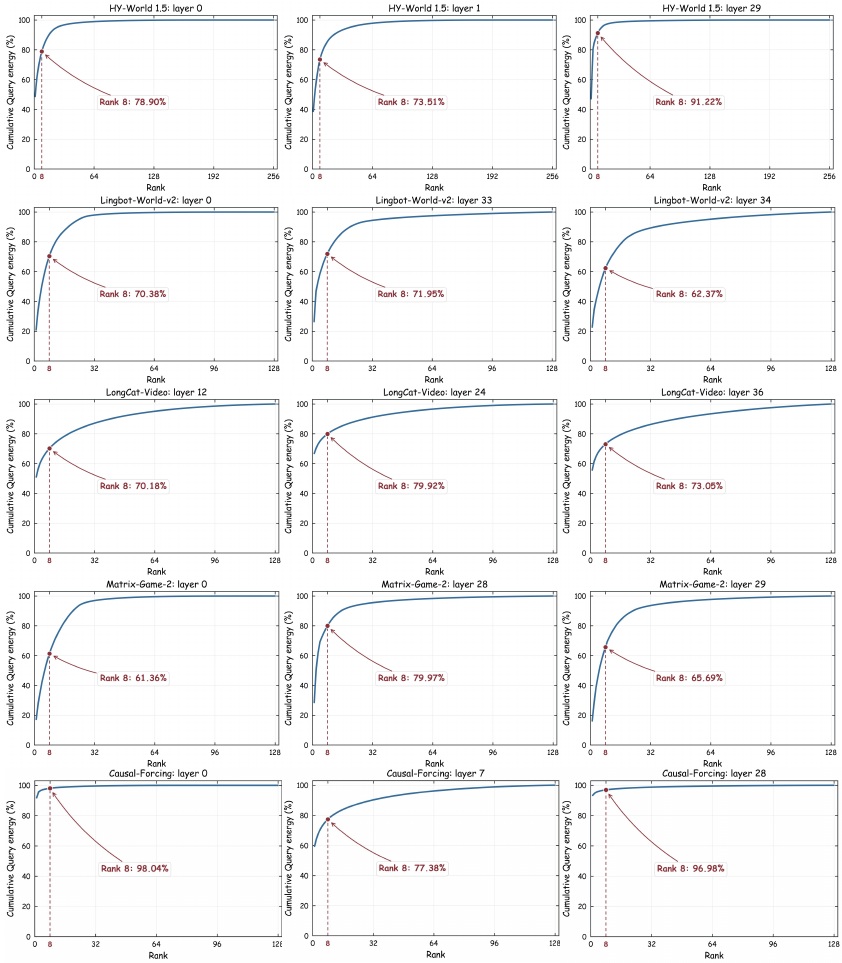}%
\caption{Cumulative Query energy across different models and layers.
}
\label{all_curves}
\end{figure*}

To further validate the low-rank structure of dominant Query directions in PSAC, we visualize the cumulative energy of historical Queries on more video world models. As shown in Figure \ref{all_curves}, the top-$8$ directions capture a large fraction of the total Query energy across different models and layers, which provides a compact principal Query subspace for correcting the Key quantization errors that have larger impacts on attention logits.

\begin{table}[t]
\centering
\caption{
Ablation study on the rank $r$ of PSAC.
Increasing the rank provides marginal quality improvements but introduces higher storage overhead, so we use $r=8$ to balance visual quality and KV cache compression.
}
\label{psac_rank}

\setlength{\tabcolsep}{4.5pt}
\renewcommand{\arraystretch}{1.08}
\small

\begin{tabular}{c|c|ccc|c}
\toprule
\cellcolor{blue!7}\textbf{Model} &
\cellcolor{blue!7}\textbf{Rank} &
\cellcolor{blue!7}\textbf{PSNR} $\uparrow$ &
\cellcolor{blue!7}\textbf{SSIM} $\uparrow$ &
\cellcolor{blue!7}\textbf{LPIPS} $\downarrow$ &
\cellcolor{blue!7}\textbf{$\Delta$ Bit vs. $r=8$} \\
\midrule

\multirow{5}{*}{HY-World 1.5}
& 2
& 18.3949
& 0.7643
& 0.1102
& $-0.1006$ \\

& 4
& 18.3973
& 0.7645
& 0.1100
& $-0.0671$ \\

& \ourcell{\textbf{8}}
& \ourcell{18.4485}
& \ourcell{0.7661}
& \ourcell{0.1082}
& \ourcell{0.0000} \\

& 16
& \textbf{18.5368}
& 0.7692
& \textbf{0.1063}
& $+0.1341$ \\

& 32
& 18.5157
& \textbf{0.7694}
& 0.1067
& $+0.4023$ \\

\midrule

\multirow{5}{*}{Causal-Forcing}
& 2
& 16.2259
& 0.7457
& 0.1569
& $-0.1926$ \\

& 4
& 16.2409
& 0.7466
& 0.1554
& $-0.1284$ \\

& \ourcell{\textbf{8}}
& \ourcell{16.2828}
& \ourcell{0.7464}
& \ourcell{0.1554}
& \ourcell{0.0000} \\

& 16
& \textbf{16.3495}
& 0.7491
& 0.1527
& $+0.2569$ \\

& 32
& 16.3462
& \textbf{0.7507}
& \textbf{0.1515}
& $+0.7706$ \\

\bottomrule
\end{tabular}
\end{table}

\subsection{The Effect of the Hyper-parameter $r$ in PSAC}
We further study the effect of the PSAC rank $r$. As shown in Table \ref{psac_rank}, increasing $r$ generally improves the frame-level quality, but the gains gradually diminish. For example, increasing the rank from $8$ to $16$ only brings modest improvements in PSNR, SSIM, and LPIPS, while introducing an additional $0.1341$ bits on HY-World 1.5 and $0.2569$ bits on Causal-Forcing relative to $r=8$. This trend is also consistent with the cumulative Query-energy distributions in Figure \ref{rank8} and \ref{all_curves}, where the additional energy captured by higher-rank directions gradually decreases. Therefore, we set $r=8$ throughout our experiments as a practical trade-off between visual quality and additional storage overhead.

\subsection{Initialization of Historical Query Statistics}

At the beginning of generation, historical Query statistics are not available. To handle this cold-start stage without introducing unreliable estimates, we adopt a simple causal initialization strategy. For the first cache chunk, QSAC assigns uniform channel sensitivity, i.e., $w_{h,c}=1$, such that centroid selection depends only on the residual quantization characteristics, while PSAC is temporarily inactive because no reliable principal Query subspace can be approximated. After the current chunk is quantized, its Query observations are incorporated into the historical second-order statistics and become available to subsequent chunks. In this way, the statistics used by QSAC and PSAC are always constructed from previously generated content, which preserve strict causality throughout inference. Since each committed chunk contributes a large number of spatial Query tokens, the historical statistics can be established rapidly after initialization without requiring an additional calibration stage or manually designed warm-up schedule.

\begin{table}[t]
\centering
\caption{
Effect of the diagonal approximation in QSAC.
}
\label{diag_ablation}

\setlength{\tabcolsep}{4.5pt}
\renewcommand{\arraystretch}{1.08}
\small

\begin{tabular}{c|c|cccc}
\toprule
\cellcolor{blue!7}\textbf{Model} &
\cellcolor{blue!7}\textbf{QSAC Metric} &
\cellcolor{blue!7}\textbf{PSNR} $\uparrow$ &
\cellcolor{blue!7}\textbf{SSIM} $\uparrow$ &
\cellcolor{blue!7}\textbf{LPIPS} $\downarrow$ &
\cellcolor{blue!7}\textbf{Latency (s)} $\downarrow$ \\
\midrule

\multirow{2}{*}{HY-World 1.5}
& \ourcell{\textbf{Diagonal}}
& \ourcell{\textbf{18.4485}}
& \ourcell{0.7661}
& \ourcell{\textbf{0.1082}}
& \ourcell{\textbf{118.75}} \\

& Full
& 18.3827
& \textbf{0.7665}
& 0.1097
& 123.10 \\

\midrule

\multirow{2}{*}{Causal-Forcing}
& \ourcell{\textbf{Diagonal}}
& \ourcell{\textbf{16.2828}}
& \ourcell{\textbf{0.7464}}
& \ourcell{0.1554}
& \ourcell{\textbf{101.21}} \\

& Full
& 16.2759
& 0.7449
& \textbf{0.1547}
& 104.07 \\

\bottomrule
\end{tabular}
\end{table}

\subsection{The Effect of the Diagonal Query Approximation in QSAC}

QSAC maintains the historical Query second-moment matrix $\mathbf{M}_h^{t-1}$ in Eq. \ref{eq6}, and uses its diagonal entries to construct the Query-aware distance in Eq. \ref{eq9}. To study the effect of the ignored cross-channel correlations, we replace Eq. \ref{eq9} with the full quadratic form:
\begin{equation}
d_{\mathrm{full}}(\mathbf{k}_i,\boldsymbol{\mu}_j)
=
(\mathbf{k}_i-\boldsymbol{\mu}_j)^\top
\mathbf{M}_h^{t-1}
(\mathbf{k}_i-\boldsymbol{\mu}_j),
\end{equation}
which is used to select the top-$m$ candidate centroids and the remaining QuantWM pipeline is unchanged. As shown in Table \ref{diag_ablation}, retaining the full matrix does not provide consistent improvements in frame-level quality. The differences in PSNR, SSIM and LPIPS are marginal on both HY-World 1.5 and Causal-Forcing, while the full matrix introduces additional inference latency. Therefore, we adopt the diagonal approximation in QSAC throughout our experiments.

\subsection{The Effect of the Number of Candidate Centroids in QSAC}
QSAC first selects the top-$M$ candidate centroids using the Query-aware distance and then determines the final centroid according to the quantization-aware score. We study the effect of $M$ on HY-World 1.5 and Causal-Forcing in Table \ref{qsac_candidate}. Increasing $M$ beyond 4 brings only marginal changes in frame-level quality. In particular, evaluating all 256 centroids does not improve PSNR, SSIM, or LPIPS over $M=4$ on either model, while significant increasing the inference latency. These results indicate that a small candidate set is sufficient to retain high-quality centroids for the subsequent quantization-aware selection, so we use $M=4$ throughout our experiments as a practical trade-off between visual quality and efficiency. Note that the latency results in Tables \ref{diag_ablation} and \ref{qsac_candidate} are mismatched because they are measured in separate cold-start runs with different GPU occupancy. Please compare them in each ablation.

\begin{table}[t]
\centering
\caption{
Effect of the number of candidate centroids $M$ in QSAC.
$M=4$ achieves a favorable trade-off between visual quality and inference cost,
while exhaustive evaluation over all 256 centroids provides no additional quality benefit.
}
\label{qsac_candidate}

\setlength{\tabcolsep}{4.0pt}
\renewcommand{\arraystretch}{1.08}
\small

\begin{tabular}{c|c|ccc|c}
\toprule
\cellcolor{blue!7}\textbf{Model} &
\cellcolor{blue!7}\textbf{$M$} &
\cellcolor{blue!7}\textbf{PSNR} $\uparrow$ &
\cellcolor{blue!7}\textbf{SSIM} $\uparrow$ &
\cellcolor{blue!7}\textbf{LPIPS} $\downarrow$ &
\cellcolor{blue!7}\textbf{Latency (s)} $\downarrow$ \\
\midrule

\multirow{7}{*}{HY-World 1.5}
& 1   & 18.2442 & 0.7612 & 0.1132 & 116.06 \\
& 2   & 18.3988 & 0.7639 & 0.1100 & 116.22 \\

& \ourcell{\textbf{4}}
& \ourcell{18.4485}
& \ourcell{0.7661}
& \ourcell{\textbf{0.1082}}
& \ourcell{118.75} \\

& 8   & 18.4393 & 0.7658 & 0.1089 & 119.51 \\
& 16  & \textbf{18.4880} & \textbf{0.7665} & 0.1086 & 120.65 \\
& 32  & 18.4701 & 0.7664 & 0.1084 & 118.97 \\
& 256 & 18.4450 & 0.7661 & 0.1094 & 240.66 \\

\midrule

\multirow{7}{*}{Causal-Forcing}
& 1   & 16.1767 & 0.7434 & 0.1567 & 135.07 \\
& 2   & 16.2433 & 0.7478 & 0.1543 & 135.84 \\

& \ourcell{\textbf{4}}
& \ourcell{16.2828}
& \ourcell{0.7464}
& \ourcell{0.1554}
& \ourcell{135.08} \\

& 8   & 16.2805 & 0.7479 & 0.1535 & 144.90 \\
& 16  & \textbf{16.2834} & \textbf{0.7487} & \textbf{0.1529} & 148.43 \\
& 32  & 16.2161 & 0.7470 & 0.1554 & 159.97 \\
& 256 & 16.2067 & 0.7463 & 0.1559 & 1386.38 \\

\bottomrule
\end{tabular}
\end{table}

\section{Efficient Algorithm-System Codesign}
\label{system}

\subsection{Implementation Details}

\begin{figure*}[t]
\centering
\includegraphics[width=5.3in]{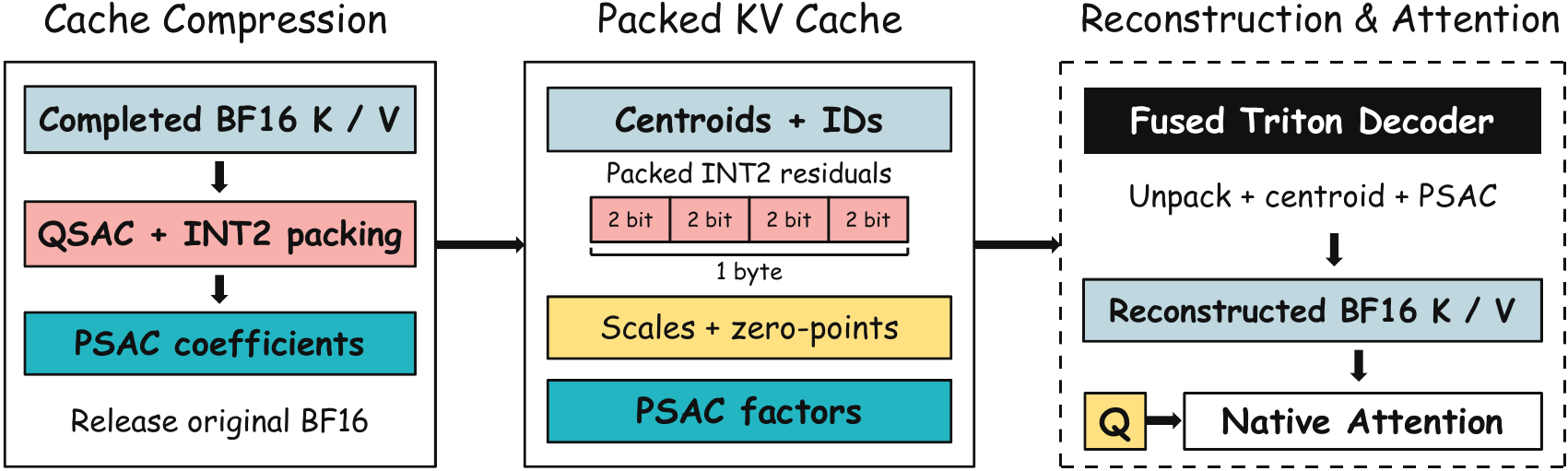}%
\caption{Overview of \textbf{Quant\textit{WM}}'s inference system. Completed KV cache chunks are stored in a packed low-bit representation. A fused decoder reconstructs BF16 keys and values on demand for native attention.
}
\label{codesign}
\end{figure*}

Figure \ref{codesign} illustrates how our \textbf{Quant\textit{WM}} integrates cache compression, packed storage and on-demand reconstruction into streaming inference.

\paragraph{Cache lifecycle}
At each native cache-chunk boundary, \textbf{Quant\textit{WM}} replaces the completed BF16 chunk with its packed representation and releases the original storage. The active chunk remains in BF16 until completion, while historical chunks remain compressed between attention calls. To support online QSAC and PSAC, we incrementally maintain per-head Query statistics and compute the principal basis at chunk boundaries without retaining historical Query tensors.

\paragraph{Packed storage}
We physically pack four INT2 residual codes into each byte and store centroid assignments as \texttt{uint8} indices. The zero-points are also bit-packed as INT2, while group scales use INT8 codes with a shared FP16 secondary scale. Each INT8 PSAC coefficient row has an FP16 scale. We retain a quantized correction only when it reduces the approximated Query-weighted error; otherwise, its coefficient row is set to zero. For each cache chunk, we select the smaller of dense and sparse coefficient layouts, including the storage cost of sparse indices. 

\paragraph{Fused reconstruction}
During each KV-cache access for attention computation, a Triton kernel fuses residual unpacking, quantization-parameter reconstruction, centroid gathering and PSAC accumulation. Values follow the same reconstruction path without PSAC. The kernel writes only the reconstructed BF16 K/V, which avoids separate dense intermediates for centroid addition and low-rank compensation. The model's native FlashAttention or SDPA implementation consumes these temporary tensors, which are not retained in the persistent cache. We preserve each model's native cache layout, attention mask and RoPE convention, where pre-RoPE Keys receive positional encoding after reconstruction, while post-RoPE Keys are reconstructed directly.

\subsection{The Impact of KV Cache Quantization on Latency is Model-dependent}
Although KV cache quantization is mainly designed to reduce memory consumption, its impact on end-to-end latency can vary across different system configurations. Online quantization and cache reconstruction introduce additional computation, while the compressed representation reduces the amount of KV data accessed and transferred during generation. For models that are more constrained by KV-cache memory traffic or cache offloading, the reduction in data movement can outweigh the additional quantization overhead and lead to lower latency. LongCat-Video in Table \ref{latency} is such a case, where its native inference pipeline offloads a large KV cache, and QuantWM reduces the amount of cache data repeatedly transferred between host and GPU, which brings lower end-to-end latency. In contrast, when the KV cache is primarily GPU-resident, the reduction in data movement is smaller and the additional quantization operations may introduce a modest latency overhead. Therefore, the latency impact of KV cache quantization depends not only on the compression ratio, but also on how cached representations are accessed and moved during inference.

\end{document}